\documentclass[letterpaper, 10 pt, conference]{ieeeconf}  
\usepackage[dvipsnames,table]{xcolor}
\usepackage{adjustbox}
\usepackage{colortbl}
\usepackage{array}
\usepackage{url}
\usepackage{multicol,multirow}
\usepackage{booktabs}
\usepackage{subcaption}
\usepackage{float}
\usepackage{amsmath}
\usepackage{xspace}
\usepackage{graphicx}
\usepackage{tabularx}
\usepackage{multirow}
\usepackage{rotating}
\usepackage{array} 
\let\labelindent\relax
\usepackage{enumitem}
\definecolor{cvprblue}{rgb}{0.21,0.49,0.74}
 \usepackage[pagebackref,breaklinks,colorlinks,allcolors=cvprblue]{hyperref}
\usepackage{pifont}
\usepackage{amssymb}
\usepackage{tikz}

\usepackage{booktabs,multirow,array}

\newcolumntype{Y}{>{\centering\arraybackslash\bfseries}X}

\IEEEoverridecommandlockouts                              

\newcolumntype{H}{>{\setbox0=\hbox\bgroup}c<{\egroup}@{}}

\title{\LARGE \bf
SLAT-Phys: Fast Material Property Field Prediction from Structured 3D Latents}

\author{Rocktim Jyoti Das$^{1}$, Dinesh Manocha$^{1}$%
\thanks{$^{1}$University of Maryland, College Park}%
\thanks{email: rocktimj@umd.edu}%
\thanks{Website: {\footnotesize \url{https://rocktimjyotidas.github.io/projects/SLAT-Phys/}}}
}

\newcommand{\framework}{{{SLAT-Phys}}\xspace}

\begin{document}

\maketitle
\thispagestyle{empty}
\pagestyle{empty}

\begin{abstract}
Estimating the material property field of 3D assets is critical for physics-based simulation, robotics, and digital twin generation. Existing vision-based approaches are either too expensive and slow or rely on 3D information. We present \framework, an end-to-end method that predicts spatially varying material property fields of 3D assets directly from a single RGB image without explicit 3D reconstruction. Our approach leverages spatially organised latent features from a pretrained 3D asset generation model that encodes rich geometry and semantic prior, and trains a lightweight neural decoder to estimate Young's modulus, density, and Poisson's ratio. The coarse volumetric layout and semantic cues of the latent representation about object geometry and appearance enable accurate material estimation. Our experiments demonstrate that our method provides competitive accuracy in predicting continuous material parameters when compared against prior approaches, while significantly reducing computation time. In particular, SLAT-Phys requires only 9.9 seconds per object on an NVIDIA RTXA5000 GPU and avoids reconstruction and voxelization preprocessing. This results in 120x speedup compared to prior methods and enables faster material property estimation from a single image.

\end{abstract}

\section{Introduction}

Understanding the physical properties of objects from visual observations is a long-standing challenge in computer vision, physically-based simulation, and robotics~\cite{Adelson2001OnSS, Standley2017image2massET, Bell2014MaterialRI, Lin2018LearningML, Zhai2024PhysicalPU, Wu2016Physics1L,Zhang2024PhysDreamerPI, Le2025PixieFA, Dagli2025VoMPPV,Zhang2025ParticleGridND, Pumarola2020DNeRFNR}. Accurate estimation of spatially varying mechanical properties, such as Young’s modulus$(E)$, density$(\rho)$, and Poisson’s ratio$(\nu)$ is important for realistic deformation modeling, stable robotic manipulation, contact-rich interaction, and high-fidelity digital twin construction. In some robotics applications, incorrect material assumptions can lead to unstable grasps or inaccurate force prediction. In physics-based simulation and virtual environments, physically implausible material parameters result in unrealistic animations and non-physical behavior. Consequently, material property estimation~\cite{Jiang2025PhysTwinPR, Zhang2024PhysDreamerPI} has become a critical component in the construction of 3D representations that are simulation-ready of real-world objects.

\begin{figure}
    \centering
    \includegraphics[
width=0.78\linewidth,
trim=0cm 8cm 24cm 0cm,
clip
]{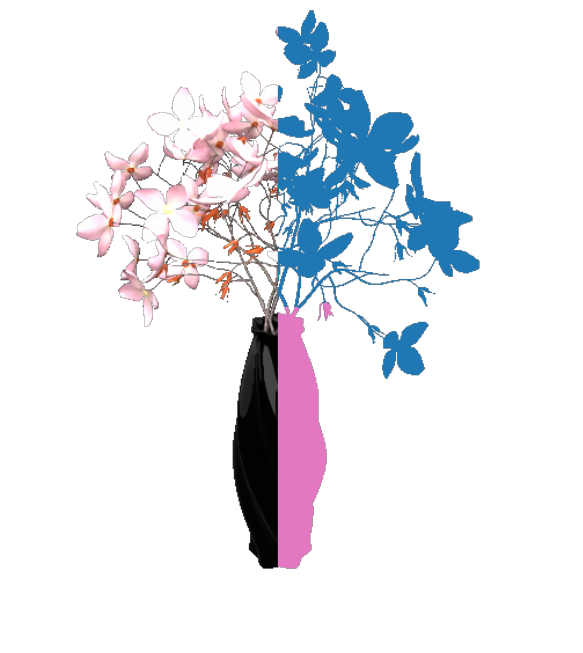}
    \caption{Different material field for different parts of the flower vase. The flower leaves are deformable whereas the vase is rigid.}
    \label{fig:placeholder}
\end{figure}
Current 3D capture methods and widely-used 3D datasets~\cite{Dong2025DigitalTC, Deitke2022ObjaverseAU, Deitke2023ObjaverseXLAU} cover a wide range of 3D geometric shapes and mesh representations, but rarely contain any annotation of physical properties. This forces designers and simulation engineers to manually annotate or estimate the material properties, which can be inaccurate, expensive, and also leads to problems like subjectivity. Recently, there has been a lot of interest in developing algorithms~\cite{Zhang2024PhysDreamerPI, Huang2024DreamPhysicsLP, Zhai2024PhysicalPU, Le2025PixieFA, Dagli2025VoMPPV} for physical property annotation of objects. Methods such as~\cite{Zhang2024PhysDreamerPI, Jatavallabhula2021gradSimDS, Huang2024DreamPhysicsLP, Li2023PACNeRFPA, Zhang2025ParticleGridND, Jiang2025PhysTwinPR, Lin2025OmniPhysGS3C} employ differentiable physics solvers to iteratively optimize material parameters by matching simulated dynamics to observed signals or realism scores from generative models~\cite{Blattmann2023StableVD, Bruce2024GenieGI}.  However, the supervision signal is sparse for predicting
physical parameters for hundreds of thousands of particles of MPM Simulation~\cite{Xie2023PhysGaussianP3, Zhang2024PhysDreamerPI}  and is an extremely slow and difficult optimization process, often taking hours on a single scene.

In parallel, there have been efforts to develop vision based pipelines~\cite{Zhai2024PhysicalPU, Le2025PixieFA, Dagli2025VoMPPV,chopra2025physgs} for physical property estimation, using semantic representation learned by 2D vision foundation models~\cite{Oquab2023DINOv2LR, Radford2021LearningTV}. These methods reconstruct the scene from multi-view images using 3D reconstruction techniques such as NeRF~\cite{mildenhall2021nerf} or Gaussian Splatting~\cite{Kerbl20233DGS}. Subsequently, a 3D semantic feature field is constructed by projecting dense 2D image features onto the resulting voxels from the reconstruction. While effective, such pipelines are computationally expensive require approximately 20 minutes of processing time on a NVIDIA RTXA5000 GPU for a single object as shown by our evaluation. The 3D reconstruction and feature extraction stages significantly increase inference time and limit scalability, especially when real-time material inference is required for interactive robotics tasks or large-scale dataset processing. Many robotic applications require real-time understanding of physical properties to grasp objects without slipping, determine appropriate interaction forces~\cite{Gao2023TheOF}, and estimate terrain traversability for navigation in outdoor environments~\cite{Seneviratne2024CROSSGAiTCM, weerakoon2023graspe, Elnoor2025VLMGroNavRN}. Furthermore, fast single-view estimation of physical properties, combined with single-view 3D asset generation~\cite{Xiang2024Structured3L, Yang2023SAM3DSA}, can accelerate both Real2Sim~\cite{Xie2025Vid2SimRA, Escontrela2025GaussGymAO} and Sim2Real~\cite{Rudin2021LearningTW} paradigms in robotics.


In this paper, we explore the use of large-scale 3D asset generation models to estimate the material properties.
Large-scale 3D asset generation models~\cite{Xiang2024Structured3L, Yang2023SAM3DSA} have been useful for single-image 3D understanding. Trained on massive datasets~\cite{Deitke2022ObjaverseAU, Deitke2023ObjaverseXLAU}, these models learn structured latent representations that encode rich geometric and semantic priors. In particular, Structured LATent (SLAT)~\cite{Xiang2024Structured3L} representations which are structured latent features organized spatially in 3D, enabling high-quality 3D reconstruction from a single RGB image. These structured latents implicitly capture multi-view consistency, volumetric occupancy cues, and object-level semantics without requiring explicit test-time multi-view aggregation.  
Our goal is to leverage these pretrained generative priors for physically grounded inference, and thereby reduce the computational cost.

\noindent{\bf Main Results:}
In this work, we introduce \textbf{\framework}, an end-to-end feedforward network that directly predicts continuous, spatially varying material property fields from structured 3D latent representations. Given a single RGB image, we extract spatially organized SLAT features from a pretrained 3D generation backbone. Unlike prior work which require constructing explicit geometry~\cite{Zhai2024PhysicalPU,Le2025PixieFA}, we utilize the semantic and geometric information encoded in the structured latent space and employ a lightweight neural decoder to regress Young’s modulus $(E)$, density$(\rho)$, and Poisson’s ratio$(\nu)$.  Operating purely in latent space yields two major advantages. First, it removes expensive reconstruction stages such as multi-view rendering, voxelization, and 3D semantic feature construction, leading to much faster inference compared to reconstruction-based pipelines. Second, it allows us to test a broader hypothesis about representation learning: that structured generative latents encode sufficient physically grounded information to support downstream physical reasoning tasks.

Extensive experiments conducted under the Pixie~\cite{Le2025PixieFA} evaluation protocol demonstrate that \framework achieves competitive accuracy in material prediction while reducing the inference time by $\sim 120$ times. Our results indicate that structured 3D latents are not merely geometric priors for reconstruction but also encode physically informative signals that can be harnessed for simulation-ready digital twin generation from a single image. By bridging large-scale 3D generative modeling and material-aware physical inference, SLAT-Phys opens a new direction toward real-time, physically grounded 3D understanding. Rather than treating reconstruction and physics estimation as separate stages, our work suggests that structured latent representations can serve as a unified foundation for both geometry and physical parameter reasoning.

Our key contributions are summarized as follows:

\begin{itemize}
    \item We introduce \textbf{\framework}, the first approach to directly regress continuous material property fields from a single image without explicit 3D reconstruction or multi-view aggregation.
    
    \item We demonstrate that spatially organized SLAT features learned by large-scale 3D generation models encode physically meaningful information beyond geometry and appearance, enabling accurate estimation of Young’s modulus $(E)$, density $(\rho)$, and Poisson’s ratio $(\nu)$.
    
    
    \item Through extensive evaluation, we show that SLAT-Phys achieves competitive material prediction accuracy while enabling faster, simulation-ready digital twin generation from a single RGB image, achieving \textbf{120$\times$ faster inference}.
\end{itemize}
\section{Related Work}

\subsection{Vision-based Physics Estimation}

Inferring physical properties of objects from visual observations is a long-standing problem~\cite{Adelson2001OnSS, Wu2016Physics1L, Bell2014MaterialRI, Lin2018LearningML}. Early work such as \textit{image2mass}~\cite{Standley2017image2massET} demonstrated that object mass can be estimated from a single image by learning correlations between appearance, scale, and category priors. However, these approaches typically predict only global object-level properties and cannot recover spatially varying material fields. NeRF2Physics~\cite{Zhai2024PhysicalPU} leverages vision–language models and neural feature fields to infer material properties by reasoning over candidate materials suggested by language priors. Along similar lines, PhysGS~\cite{chopra2025physgs} is combining Gaussian Splatting~\cite{Kerbl20233DGS} with Bayesian inference~\cite{robert2010bayesianinference} to give get prediction with an uncertainity estimate. Other approaches such as PhysDreamer~\cite{Zhang2024PhysDreamerPI}, PGND~\cite{Zhang2025ParticleGridND} and PhysTwin~\cite{Jiang2025PhysTwinPR} estimate physics parameters by matching simulated dynamics to observations using differentiable simulation or generative video models. While effective, these methods often require expensive per-object optimization and may produce simulator-specific parameters that do not generalize across different physics engines.

A recent line of work~\cite{Dagli2025VoMPPV, Le2025PixieFA} instead learns feed-forward mappings from visual or 3D features to spatially varying mechanical properties. The volumetric fields of Young’s modulus, density, and Poisson’s ratio is predicted from 3D representations using learned volumetric architectures. However, these methods typically rely on explicit 3D reconstructions or multi-view feature aggregation to obtain volumetric representations. In contrast, SLAT-Phys operates directly in the structured latent space of a pretrained 3D generative model~\cite{Xiang2024Structured3L, Yang2023SAM3DSA} and predicts spatially varying mechanical properties from a single image without explicit 3D reconstruction, enabling significantly faster inference while maintaining competitive accuracy.

\subsection{Learned 3D Representations}

Traditional 3D representations such as meshes, voxel grids, and signed distance fields (SDFs) explicitly encode object geometry but do not capture higher-level semantic or material information, limiting their usefulness for reasoning about physical properties. Early work~\cite{Lin2018LearningML} proposed learning material-aware local descriptors on 3D surfaces, where a projective CNN extracts view-based features around surface points to classify material labels, demonstrating that local geometric context can provide cues about object materials. More recent work such as F3RM~\cite{Shen2023DistilledFF} has focused on incorporating semantic information into spatial representations by aggregating features from pretrained vision models~\cite{Radford2021LearningTV,Oquab2023DINOv2LR}. This multi-view feature aggregation paradigm is also adopted by recent physics estimation frameworks such as NeRF2Physics~\cite{Zhai2024PhysicalPU} and Pixie~\cite{Le2025PixieFA}, which rely on explicit 3D reconstruction and multi-view feature lifting to estimate material properties. However, these approaches require expensive reconstruction pipelines and multi-view aggregation. They take approximately 20 mins of inference time per object on an NVIDIA RTXA5000 GPU. More recently, large-scale 3D generation models such as  TRELLIS~\cite{Xiang2024Structured3L, Yang2023SAM3DSA} proposes Structured LATents (SLAT), a spatially organized latent representation that encode geometric and semantic priors directly from data. In this work, we build on this representation and developed \textbf{\framework} to directly predict spatially varying material property fields from a single image without explicit 3D reconstruction.

\begin{figure*}[t]
\centering
\includegraphics[
width=1\linewidth,
trim=0cm 8cm 4cm 0cm,
clip
]{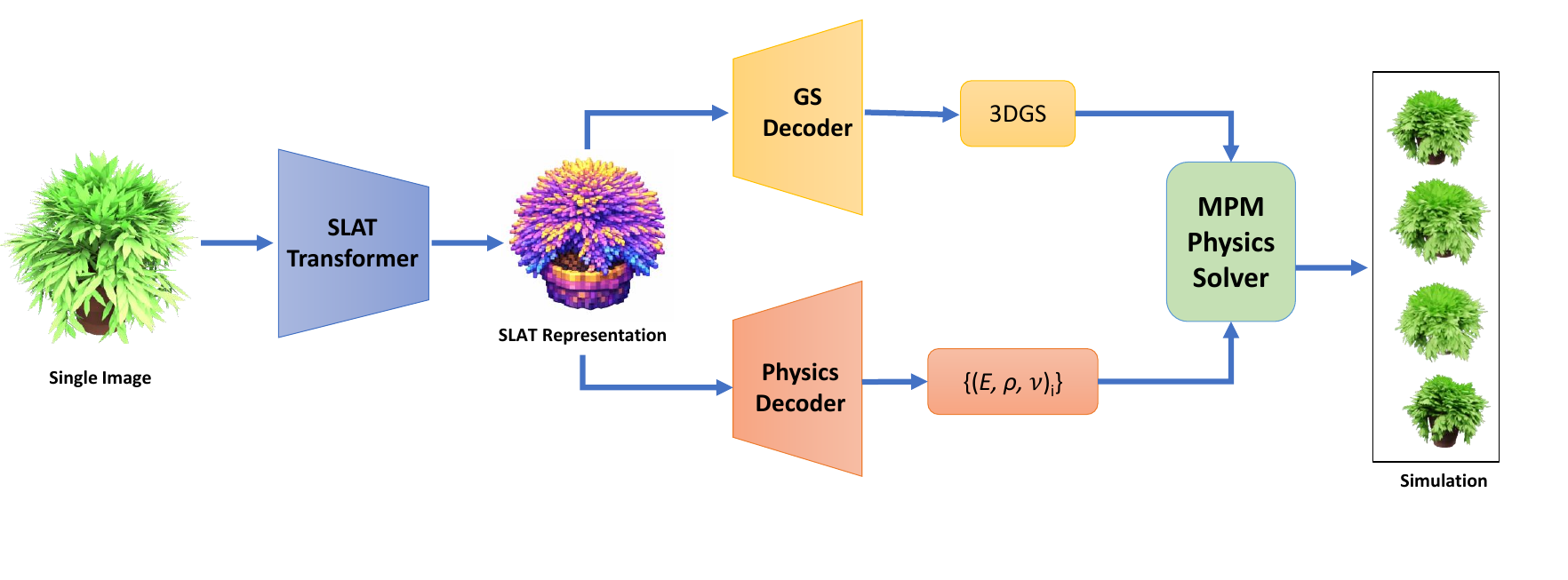} 
\caption{\textbf{Overview of \framework:} Given a single image of a 3D asset, our framework first encodes the image using the Trellis encoder to obtain a structured 3D latent (SLAT) representation. From this latent, two decoders predict the corresponding 3D Gaussian Splatting (3DGS) representation and spatially varying material properties $(E, \rho, \nu)$ at each voxel. The resulting Gaussian representation together with the predicted physical parameters are then passed to an MPM solver to perform physics-based simulation.}
\label{fig:pipeline}
\end{figure*}
\section{Our Method: SLAT-Phys}

\subsection{Method Overview}

Our goal is to estimate spatially varying material properties of a 3D object from a single RGB image. Specifically, we predict per-voxel mechanical parameters including Young's modulus $E$, density $\rho$, and Poisson's ratio $\nu$, along with a discrete material label. Given an accurate and valid triplet $(E, \nu, \rho)$ along with a reasonable material model, a consistent numerical simulation such as MPM~\cite{Hu2018AML} can
produce accurate predictions of an object’s behavior under external force.

Our approach consists of three stages. First, a pretrained image-to-3D generation model extracts a sparse structured latent representation of the object from a single image. These structured latents, referred to as \textit{Structured LATents (SLAT)}, encode geometric and semantic information at a set of occupied voxel locations. Second, a lightweight neural decoder predicts per-voxel physical properties directly from the SLAT features using a sparse transformer architecture. Finally, the predicted material fields are passed to a downstream physics simulator to generate physically plausible object behavior.

\subsection{Problem Formulation}

We consider the problem of estimating spatially varying physical material properties of a 3D object from a single RGB observation. Let $I \in \mathbb{R}^{H \times W \times 3}$ denote an input RGB image containing an object. Our goal is to estimate a volumetric field of physical parameters defined over a discretized voxel grid. We represent the object using a voxel grid
\begin{equation}
\mathcal{V} = \{v_i\}_{i=1}^{N}, \quad v_i \in \mathbb{R}^3,
\end{equation}
where each voxel corresponds to a spatial location in a canonical $64^3$ grid. For each voxel we aim to predict a vector of physical parameters
\begin{equation}
\mathbf{p}_i =
\left[
E_i, \rho_i, \nu_i, c_i
\right],
\end{equation}
where $E_i$ denotes Young's modulus, $\rho_i$ the density, $\nu_i$ the Poisson's ratio, and $c_i$ a discrete material class label. Instead of predicting these parameters from dense 3D geometry, we leverage a structured latent representation produced by a pretrained 3D generative model. Given an input image $I$, the encoder of the 3D generation model produces a set of sparse latent features

\begin{equation}
\mathcal{Z} =
\{(\mathbf{x}_i, \mathbf{z}_i)\}_{i=1}^{N},
\end{equation}
where $\mathbf{x}_i$ denotes the voxel coordinate and $\mathbf{z}_i \in \mathbb{R}^d$ denotes the latent feature vector. Our model learns a mapping
\begin{equation}
f_\theta : (\mathbf{x}_i, \mathbf{z}_i) \rightarrow
(E_i, \rho_i, \nu_i, c_i),
\end{equation}
where $f_\theta$ is implemented as a sparse transformer decoder operating on the structured latent representation.

\subsection{SLAT Feature Extraction}

Given an RGB image $I$, we use a pretrained image-to-3D model to extract a structured latent representation of the object. Our encoder predicts a sparse set of occupied voxel coordinates and associated latent features. For each object the model produces a set
\begin{equation}
\{(\mathbf{x}_i, \mathbf{z}_i)\}_{i=1}^{N},
\end{equation}
where $\mathbf{x}_i \in \{0,\dots,63\}^3$ denotes the voxel coordinate and $\mathbf{z}_i \in \mathbb{R}^{8}$ represents an 8-dimensional latent feature vector. These voxels correspond to the predicted surface of the reconstructed object and form a sparse representation of the geometry. The TRELLIS encoder is kept frozen during training, allowing the model to benefit from geometric priors learned from large-scale 3D generative training.

\subsection{Physics Decoder}
\label{sec:physics_decoder}
The structured latent features from the pretrained image-to-3D model are processed by a neural decoder that predicts physical properties for each voxel. The decoder is implemented as a sparse Swin-style transformer~\cite{Liu2021SwinTH} operating directly on the occupied voxel set. The network consists of an input projection layer, positional encoding, four sparse transformer blocks, and regression and classification heads. Each transformer block performs windowed self-attention over local voxel neighborhoods to capture spatial dependencies between nearby latent features. The final predictions are

\begin{equation}
\hat{p}_i =
(\hat{E}_i, \hat{\rho}_i, \hat{\nu}_i, \hat{c}_i),
\end{equation}
computed using two output heads. The regression head predicts continuous physical parameters, while the classification head predicts discrete material labels.





\subsection{Training Data}
We train our model using the PixieVerse dataset introduced in Pixie~\cite{Le2025PixieFA}. 
PixieVerse is a large-scale dataset of 3D objects paired with spatially varying physical material annotations. 
The dataset contains 1,624 high-quality single-object assets spanning 10 semantic categories, including organic objects (trees, shrubs, flowers), deformable toys (e.g., rubber ducks), sports equipment (balls), granular materials (sand, snow, mud), and hollow containers (e.g., soda cans and metal crates). 
The objects are sourced from the Objaverse repository~\cite{Deitke2022ObjaverseAU, Deitke2023ObjaverseXLAU} and curated through a filtering pipeline to ensure geometric and visual quality. 
Each asset is annotated with both discrete material categories and continuous mechanical parameters, including Young’s modulus $(E)$, Poisson’s ratio $(\nu)$, and mass density $(\rho)$, enabling simulation-ready material fields. 
The annotations are generated through a semi-automatic labeling pipeline that leverages vision-language models~\cite{Comanici2025Gemini2P} and CLIP feature fields~\cite{Radford2021LearningTV}, combined with manual verification to ensure physical plausibility. 
These annotations provide voxel-level supervision for learning spatially varying material property fields from visual observations. 
\label{sec:training_data_generation}

\subsection{Training Objective}
\label{sec:train_obj}
Following Pixie~\cite{Le2025PixieFA}, our model is trained to jointly predict continuous physical parameters and material classes. For each voxel with valid annotation we compute regression losses between the predicted and ground truth per voxel continuous physics:
\begin{align}
\mathcal{L}_E &= \| \hat{E}_i - E_i \|_2^2, \\
\mathcal{L}_\rho &= \| \hat{\rho}_i - \rho_i \|_2^2, \\
\mathcal{L}_\nu &= \| \hat{\nu}_i - \nu_i \|_2^2,
\end{align}
along with a material classification loss

\begin{equation}
\mathcal{L}_{mat} =
\text{CE}(\hat{c}_i, c_i),
\end{equation}
where CE denotes cross-entropy loss. The final training objective is
\begin{equation}
\mathcal{L} =
\lambda_E \mathcal{L}_E +
\lambda_\rho \mathcal{L}_\rho +
\lambda_\nu \mathcal{L}_\nu +
\lambda_{mat}\mathcal{L}_{mat}.
\end{equation}





\section{Experiments}

Our experiments are designed to assess the performance of \framework in terms of  performance and runtime speed.


\subsection{Implementation Details}

\paragraph{SLAT Generation}
For extracting structured latent features, we employ the TRELLIS image-to-3D generation framework~\cite{Xiang2024Structured3L}. 
Given a single RGB image of an object, the model produces a Structured LATent (SLAT) representation consisting of voxel coordinates and an 8-dimensional latent feature for each occupied voxel on a $64^3$ grid. 
The coordinates indicate the spatial position of voxels, while the latent vectors capture geometric and semantic cues learned by the pretrained TRELLIS model. Throughout our pipeline, the TRELLIS encoder is kept frozen and used purely as a feature extractor.

\paragraph{Dataset}
Training targets are obtained from the PixieVerse dataset introduced in Pixie~\cite{Le2025PixieFA}, which provides volumetric predictions of physical material properties on a $64^3$ voxel grid. 
For each object, PixieVerse provides normalized predictions of density $(\rho)$, Young's modulus $(E)$, and Poisson's ratio $(\nu)$, along with an 8-class material label represented as one-hot channels. The physical parameters are stored in normalized form within the range $[-1,1]$, where density and Young’s modulus are encoded in log-space and Poisson’s ratio is encoded linearly.  These voxel-level annotations serve as the supervision signal for training the physics decoder (as described in Sections~\ref{sec:physics_decoder} and ~\ref{sec:train_obj}) that maps SLAT features to material property fields.

\paragraph{Physics-to-SLAT Annotation Alignment}

\begin{figure}[t]
    \centering
    \includegraphics[width=1\linewidth]{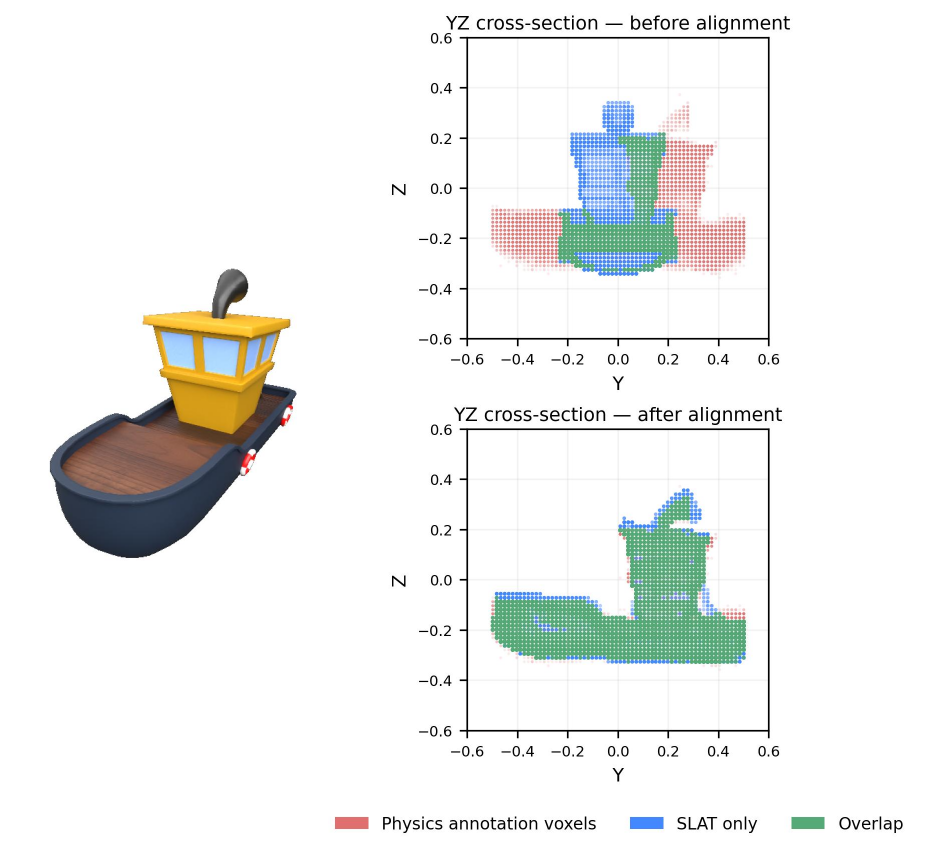}
    \caption{\textbf{Physics-to-SLAT voxel alignment.} 
    Physics annotations predicted by Pixie and the SLAT voxel grid produced by TRELLIS may exhibit a rigid rotational offset due to independent reconstruction pipelines. 
    The figure illustrates the voxel grids before alignment and after applying the estimated rigid transformation using ICP.}
    \label{fig:voxel_alignment}
\end{figure}

The physics annotations used for supervision originate from the Pixie~\cite{Le2025PixieFA}, while the SLAT representation is generated independently by the TRELLIS image-to-3D model. Although both representations share the same $64^3$ voxel resolution, they may differ by a rigid rotational offset due to the independent reconstruction pipelines. As illustrated in Fig.~\ref{fig:voxel_alignment}, we align the physics voxel grid to the SLAT coordinate frame before training.

We compute the alignment using the Iterative Closest Point (ICP) algorithm. The occupied SLAT voxels serve as the reference surface, while the boundary voxels of the Pixie physics occupancy mask are used as the source point cloud. To avoid poor local minima, we evaluate 64 candidate orientations generated from combinations of $\{0^\circ,90^\circ,180^\circ,270^\circ\}$ rotations around the three axes and select the initialization with the highest ICP fitness score. A final ICP optimization then estimates a rigid transformation $T \in SE(3)$ that maps the physics annotations into the SLAT coordinate frame. The aligned material properties $(E,\rho,\nu)$ are subsequently queried at the corresponding SLAT voxel locations to provide supervision during training.

\paragraph{Decoder Architecture}
Following the 3D decoder design of TRELLIS~\cite{Xiang2024Structured3L}, we implement a lightweight sparse Swin-style transformer~\cite{Liu2021SwinTH} operating on the SLAT voxels. Each occupied voxel is represented by its coordinate in a $64^3$ grid and an 8-dimensional latent feature vector. The latent features are projected to a 256-dimensional embedding and processed by eight transformer blocks with sixteen attention heads. The resulting voxel features are passed to two prediction heads: a regression head that predicts the normalized physical parameters $(E,\rho,\nu)$ using a $\tanh$ activation, and an auxiliary classification head that predicts one of eight material classes.

\paragraph{Training Details}
The model is trained using a combination of regression and classification losses. Mean squared error (MSE) losses are applied to the normalized physical parameters $(E, \rho, \nu)$, while a cross-entropy loss supervises the auxiliary material classification task. The total loss is defined as a weighted sum of these components, with weights $1.0$ for each regression term and $0.5$ for the classification loss. Training is performed using the AdamW optimizer~\cite{Loshchilov2017DecoupledWD} with a learning rate of $10^{-4}$ and cosine annealing scheduling. We employ mixed precision training and gradient accumulation to improve training efficiency.

\begin{figure*}[t]
\centering
\includegraphics[
width=1\linewidth,
trim=3cm 3cm 7cm 1cm,
clip
]{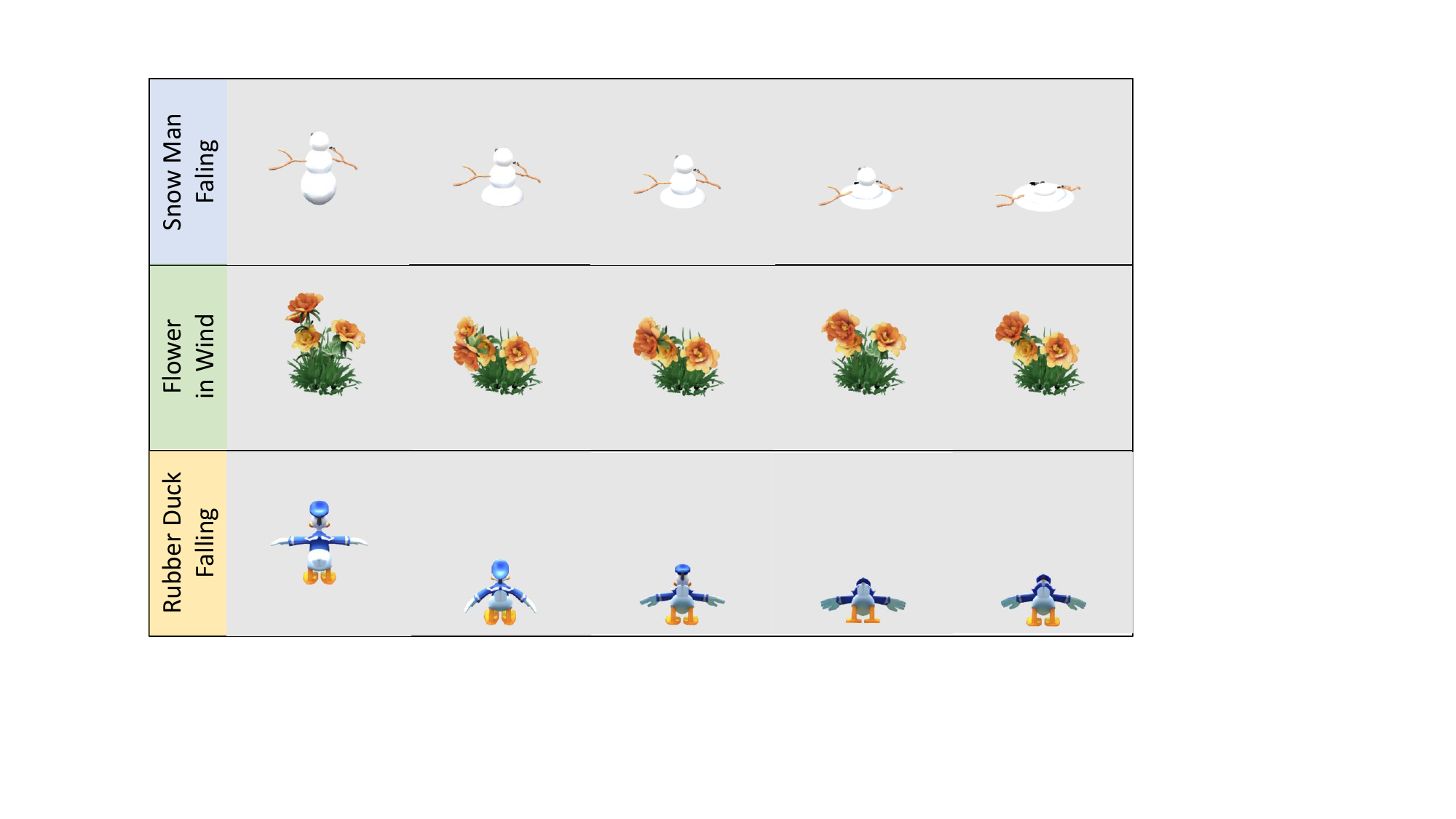}
\caption{Physics-based simulation results using the predicted physical parameters computed using \framework. 
Given a single image, our method predicts spatially varying Young's modulus ($E$), Poisson's ratio ($\nu$), and density ($\rho$) from SLAT features. The same SLAT representation is also decoded to obtain 3D Gaussian splats for geometry and appearance reconstruction. 
The resulting geometry and predicted physical parameters are then used in an MPM simulator to generate physically plausible dynamics. 
We highlight the simulation results for three objects with different material behaviors: a snow man and a rubber duck falling under gravity and a flower under the influence of wind.}
\label{fig:similation}
\end{figure*}

\paragraph{Physics-Based Simulation}
We evaluate the predicted material fields using the Material Point Method (MPM), a particle–grid hybrid approach for simulating deformable materials~\cite{Xie2023PhysGaussianP3}. The MPM solver takes as input a set of particles representing the object geometry along with their associated material parameters and external force specifications. In our approach, both geometry and physics originate from the same SLAT representation produced by TRELLIS. A Gaussian Splatting model is decoded from the SLAT features to reconstruct the object geometry, where each Gaussian naturally corresponds to a simulation particle~\cite{Xie2023PhysGaussianP3}. The predicted material properties, Young’s modulus $(E)$, density $(\rho)$, and Poisson’s ratio $(\nu)$, are transferred from the SLAT voxel grid to the Gaussian particles via nearest-neighbor interpolation. The resulting set of particles with spatially varying material parameters is then passed to the MPM solver to simulate object deformation under external forces. We highlight the results for different models in Fig.~\ref{fig:similation}.

\subsection{Ablation Study: Physics Decoder Architecture}

\begin{table}[h]
\centering
\small
\begin{tabular}{lcccc}
\toprule
\textbf{Model} &  log $E$ err  $\downarrow$ & log $\rho$ err $\downarrow$ & $\nu$ error $\downarrow$ & Mat. Acc. $\uparrow$ \\
\midrule
Small  & 0.0285 & 0.0630 & 0.0674 & 0.9107 \\
Medium  & 0.0195 & 0.0433 & 0.0573 & 0.9355 \\
Large & \textbf{0.0173} & \textbf{0.0361} & \textbf{0.0540} & \textbf{0.9453} \\
\bottomrule
\end{tabular}
\caption{Ablation over Physics Decoder architectures of increasing capacity. All models are trained on the same data with identical optimization settings.}
\label{tab:physics_decoder_ablation}
\end{table}

We study the impact of model capacity by varying the number of channels, transformer blocks, and attention heads in the Physics Decoder. All models share the same Swin-attention-based sparse voxel architecture operating on a 64³ grid with 8 latent input channels, and are trained for 2,000 steps under identical optimization settings (AdamW, lr=1e-4, cosine annealing). The Small model (64 channels, 4 blocks, 4 heads, 0.20M parameters) serves as the baseline. The Medium model (128 channels, 6 blocks, 8 heads, 1.19M parameters) doubles the channel width and adds two additional transformer blocks, yielding roughly a 6× parameter increase. The Large model (256 channels, 8 blocks, 16 heads, 6.32M parameters) further doubles the channel width and adds two more blocks, resulting in a 31× parameter increase over Small.

The Medium and Large models require fewer steps and we selected checkpoints before model starts overfitting and the results are reported in Table~\ref{tab:physics_decoder_ablation}. We can observe that increasing model capacity from Small to Large leads to consistent improvements across all regression metrics and material classification accuracy, indicating that higher representational capacity is beneficial for modeling physical properties. The Large model provides the strongest overall performance.
\begin{table*}[h]
\centering

\begin{subtable}{\textwidth}
\centering
\small
\begin{tabular}{lccc}
\toprule
\textbf{Stage} & \textbf{NeRF2Physics} & \textbf{Pixie} & \textbf{\framework} \\
\midrule
\multirow{2}{*}{Preprocessing} & \multirow{2}{*}{Blender rendering: $\sim$65s} & \multirow{2}{*}{Blender rendering: $\sim$65s} & Image preprocessing + \\
 &  &  & DINO encoding: 0.138s \\
3D Reconstruction & NeRF training: 1020s & F3RM training: 575.3s & Sparse structure sampling: 4.69s \\
\multirow{2}{*}{Feature Extraction} & \multirow{2}{*}{CLIP feature fusion: 96.8s} & F3RM rendering + & \multirow{2}{*}{SLAT feature sampling: 5.03s} \\
 &  & Voxelization: 606.6s & \\
\multirow{2}{*}{Physics Decoding} & BLIP-2 captioning + LLM proposal + & \multirow{2}{*}{Neural inference: 14.1s} & \multirow{2}{*}{Physics decoder: 0.004s} \\
 & Property prediction: 14s &  & \\
\midrule
\textbf{Total per object} & $\sim$1196s ($\sim$20 min) & $\sim$1261s ($\sim$21 min) & \textbf{$\sim$9.9s} \\
\textbf{Speedup vs Ours} & $121\times$ & $128\times$ & 1$\times$ \\
\bottomrule
\end{tabular}
\caption{Runtime performance of our approach with prior methods, and analyze the running time in different modules. We observe almost 120X speedup over prior methods.}
\label{tab:runtime_sub}
\end{subtable}

\vspace{6pt}

\begin{subtable}{\textwidth}
\centering
\small
\begin{tabular}{lccccc}
\toprule
Method & Mat. Acc. $\uparrow$ & Avg. Cont. MSE $\downarrow$ & log $E$ err $\downarrow$ & $\nu$ error $\downarrow$ & log $\rho$ err $\downarrow$ \\
\midrule
NeRF2Physics & 0.274 $\pm$ 0.01 & 0.858 $\pm$ 0.109& 1.115 $\pm$0.165 & 0.462 $\pm$ 0.106 & 0.997 $\pm$ 0.162 \\
Pixie& \textbf{0.985}$\pm$ 0.011 &  0.056 $\pm$0.005 & 0.022$\pm$ 0.004 & \textbf{0.034} $\pm$ 0.006 & 0.112 $\pm$ 0.009 \\
\midrule
\framework (Ours) & 0.9453 $\pm$ 0.112 &   \textbf{0.036}$\pm$ 0.049 & \textbf{0.017} $\pm$ 0.025  &  0.054$\pm$ 0.074 & \textbf{0.036} $\pm$ 0.049 \\
\bottomrule
\end{tabular}
\caption{Material property prediction accuracy comparison.}
\label{tab:accuracy_sub}
\end{subtable}

\caption{Comparison with prior methods. (a) Runtime comparison across pipelines. (b) Material property prediction performance.}
\label{tab:comparison}
\end{table*}

\subsection{Quantitative Evaluation}

\paragraph{Evaluation Metrics}
Following PIXIE~\cite{Le2025PixieFA}, we evaluate the physics decoder using mean squared error (MSE)
in normalized space for each continuous material property, and classification accuracy
for the discrete material class. Specifically, Young's modulus $E$ and density $\rho$
are first log$_{10}$-transformed, then all three properties ($\log E$, $\log \rho$, $\nu$)
are linearly normalized to $[-1, 1]$ using the dataset min/max statistics.
MSE is computed per voxel, averaged per object,
then averaged globally. The aggregate continuous MSE is the mean of the three
per-property MSEs. Material accuracy is the fraction of occupied voxels whose
predicted material class matches the ground-truth label. These metrics measure the deviation between predicted and ground-truth physical properties at corresponding voxel locations. 
\paragraph{Baselines}
We compare our method with two representative prior approaches for estimating spatially varying mechanical properties from visual observations: NeRF2Physics~\cite{Zhai2024PhysicalPU}, and Pixie~\cite{Le2025PixieFA}. NeRF2Physics assigns material properties by querying a large language model (LLM) for plausible materials and propagating these values to 3D points using CLIP feature similarities. In contrast, Pixie is feedforward models~\cite{Ronneberger2015UNetCN} that predict volumetric material fields including Young’s modulus $(E)$, density $(\rho)$, and Poisson’s ratio $(\nu)$, but both rely on an additional 3D reconstruction stage to obtain geometry-aware representations before physics prediction. 

\paragraph{Mechanical Property Estimation}

Table~\ref{tab:comparison}\subref{tab:accuracy_sub}  compares \framework with prior methods on the PixieVerse dataset. Compared to NeRF2Physics~\cite{Zhai2024PhysicalPU}, our method achieves substantially better performance across all metrics. 


SLAT-Phys and Pixie~\cite{Le2025PixieFA} are comparable with each other. These results demonstrate that accurate material property estimation can be achieved directly from structured latent representations without requiring explicit multi-view reconstruction.

\paragraph{Run-Time Comparison}
Table~\ref{tab:comparison}\subref{tab:runtime_sub} compares the per-object runtime of our method with NeRF2Physics~\cite{Zhai2024PhysicalPU} and Pixie~\cite{Le2025PixieFA}. All experiments are conducted on a single NVIDIA RTX A5000 GPU. Both prior approaches are dominated by per asset optimization-based 3D reconstruction pipelines, including NeRF or F3RM training, followed by rendering and voxelization, which account for the majority of their runtime.

In contrast, our method operates directly on the structured SLAT representation produced by TRELLIS, eliminating the need for reconstruction, dense rendering, and voxelization. As a result, our approach reduces the total runtime to 9.9 seconds per object, compared to $\sim$1196 seconds for NeRF2Physics and $\sim$1261 seconds for Pixie. Overall, this yields a speedup of approximately 121$\times$ over NeRF2Physics and 128$\times$ over Pixie, while maintaining competitive accuracy in material property prediction.

\subsection{Qualitative Evaluation}



Figure~\ref{fig:similation} shows qualitative simulation results using the physical parameters predicted by \framework. We consider three scenarios: a falling snowman, a flower subjected to wind, and a falling rubber duck. Each example demonstrates distinct material behaviors. The snowman fractures and collapses upon impact, reflecting brittle and weak structural properties. Importantly, while the snow body breaks apart, the attached stick arms remain intact and rigid throughout the motion, highlighting the model’s ability to capture heterogeneous material properties within a single object. The flower responds to wind forces with smooth bending and oscillatory motion, capturing flexible dynamics. The rubber duck exhibits deformable motion during free fall, consistent with elastic, rubber-like materials.

In addition to physically plausible dynamics, the rendered assets maintain high visual fidelity. The Gaussian splats are directly derived from the SLAT representation, which is also used for material prediction, without any additional training or fine-tuning. This highlights that the structured latent representation captures sufficient geometric and semantic information to support both realistic rendering and physically consistent simulation from a single image.

\section{Conclusion and Limitations}




We presented \textbf{SLAT-Phys}, a feedforward framework for spatially varying material property field estimation from a single RGB image. Unlike prior approaches that construct volumetric 3D features through multi-view projection and voxel aggregation, our method could generate material property field directly from single image. Through experiments following the Pixie~\cite{Le2025PixieFA} evaluation protocol, we demonstrate that SLAT-Phys achieves competitive material estimation accuracy while requiring only a single image as input and reducing inference time by more than two orders of magnitude (over $120\times$ faster than prior reconstruction-based pipelines). Such fast estimation of physical properties is important for many robotic applications, including adjusting grasp force to prevent slipping, reasoning about interaction forces during manipulation~\cite{Gao2023TheOF}, and estimating terrain traversability in outdoor navigation~\cite{Seneviratne2024CROSSGAiTCM, weerakoon2023graspe, Elnoor2025VLMGroNavRN}. Moreover, combining fast single-view physical property inference with single-view 3D asset generation~\cite{Xiang2024Structured3L, Yang2023SAM3DSA} can further enable scalable Real-to-Sim~\cite{Xie2025Vid2SimRA, Escontrela2025GaussGymAO} and Sim-to-Real~\cite{Rudin2021LearningTW} pipelines in robotics.

Despite these promising results, several limitations remain. First, the material annotations from Pixie~\cite{Le2025PixieFA} used for training are generated through a vision-language model (VLM) based pipeline, which require extensive prompt engineering and prompts are designed for specific objects to get reliable annotation.  In contrast, recent work such as VoMP~\cite{Dagli2025VoMPPV} adopts a more reliable annotation strategy by grounding material parameters in physically measured ranges curated from sources such as Wikipedia~\cite{ wiki_density} and engineering databases~\cite{engineering_toolbox_materials}. At the time of our experiments, the codebase and dataset of VoMP~\cite{Dagli2025VoMPPV} were not publicly available. We plan to analyze how our framework performs on that data in future work. Second, our current framework relies solely on visual input. However, vision alone may not fully capture certain physical cues related to material properties. Prior work such as ObjectFolder~\cite{Gao2023TheOF} demonstrates that multi-sensory learning, incorporating tactile and other sensing modalities, can significantly improve material understanding. Extending SLAT-Phys to incorporate additional sensory modalities such as touch or audio represents an important direction for improving robustness and physical reasoning in real-world robotic applications.

\bibliographystyle{IEEEtran}%
\bibliography{refs}

\end{document}